\documentclass{article} 
\usepackage{iclr2027_conference,times}

\usepackage{amsmath,amsfonts,bm}

\def\eqref#1{equation~\ref{#1}}

\def\1{\bm{1}}

\DeclareMathAlphabet{\mathsfit}{\encodingdefault}{\sfdefault}{m}{sl}
\SetMathAlphabet{\mathsfit}{bold}{\encodingdefault}{\sfdefault}{bx}{n}

\definecolor{darkgreen}{rgb}{0.1, 0.4, 0.0}
\definecolor{darkgrey}{rgb}{0.4, 0.3, 0.4}

\newcommand{\modelname}{Strike a Chord! Modal Kinetic Typography}
\newcommand{\modelnameshort}{Strike a Chord}

\newcommand{\phead}[1]{\par\noindent\textbf{#1}}

\usepackage{xcolor}
\usepackage{hyperref}
\usepackage{url}
\usepackage{graphicx}
\usepackage{animate}
\usepackage{float}      
\usepackage{placeins}
\usepackage{booktabs}
\usepackage{amsmath,amssymb}

\makeatletter
\renewcommand\subsubsection{\@startsection{subsubsection}{3}{\z@}%
  {-1.2ex \@plus -0.4ex \@minus -0.2ex}%
  {-0.6em}%
  {\normalfont\normalsize\bfseries}}
\makeatother

\title{\modelname}

\author{Maham Tanveer$^{1}$ \quad Jiyeon Han$^{1}$ \quad Nanxuan Zhao$^{2}$ \quad Hao Zhang$^{1}$ \\[2pt]
$^{1}$Simon Fraser University \qquad $^{2}$Adobe
}

\iclrfinalcopy   

\def\arxivbuild{}          
\def\novenuestatements{}   
\definecolor{linkblue}{HTML}{1A4D8F}
\def\projectpage{\begingroup\hypersetup{hidelinks}%
  Project page: \href{https://strikeachordkt.github.io/strikeachord/}{\textcolor{linkblue}{strikeachordkt.github.io/strikeachord}}\endgroup.}

\begin{document}

\maketitle
\lhead{}          

\begin{abstract}

We introduce \emph{modal kinetic typography}, which animates a vector glyph to express a semantic concept while keeping it legible. Our key idea is to build motion from the glyph's \emph{natural vibration modes}. Specifically, a finite-element eigenproblem assembled from the vector outline yields the glyph's softest modes, for the whole letter and for each of its parts, allowing it to bend. The problem's zero-energy solutions, i.e., rigid translations and rotations, are applied in closed form to each part, allowing parts to also move as blocks. To animate the glyph, a frozen video diffusion model supervises only the modes' amplitudes and phases. Our modal approach addresses two weaknesses of prior work. Free-form point optimization under video score distillation (SDS) moves each point and frame independently along noisy gradients, tearing the outline and causing jitter. In contrast, our modes are smooth along the outline and driven by a few whole-cycle harmonics, which restricts these gradients to smooth, seamlessly looping motion. On the other hand, structured alternatives rely on skeletons or keypoints from category-specific priors, whereas our modes come from the glyph itself; the only prior is a list naming each letter's moving parts, generated once for the whole alphabet by a language model. In modal kinetic typography, shape and motion are disentangled by construction: a single base outline is sculpted toward the concept, and the modal drive cannot alter it, so a letter can also be animated without being reshaped. Across letters and typefaces, our method produces more articulated and smoother motion than Dynamic Typography and AniClipart at comparable or better concept alignment, with less glyph tearing than Dynamic Typography, and is preferred by human raters, including in a frozen-shape setting where motion alone must carry the concept. Our results were also preferred over Astra (GPT-6) by human raters.

\ifdefined\projectpage\projectpage\fi

\end{abstract}

\section{Introduction}

\begin{figure}[!t]
\centering
\animategraphics[width=\textwidth,autoplay,loop,poster=first]{8}{figs/teaser_anim/frame_}{0}{11}
\caption{\textbf{Modal Kinetic Typography: Vibration modes of a shape decide how it can move.} A letter's vibration modes are
computed from the glyph itself. Three whole-letter modes are drawn here for visual representation. The glyph is  
separately deformed toward the prompted concept. Animating a few modes brings the letter to life while keeping it legible, with a frozen 
video diffusion model supplying the only learning signal for their amplitudes and phases. Shown examples include: a `V' as a bird
beating its wings, a `Q' as a cat flicking its tail, an unsculpted `Y' (i.e., motion-only animation without reshaping)
raising one arm. With each part of the letter carrying its own modes, the motion stays in the part the caption names, e.g., the tail 
swings while the bowl stays still. The outline bends smoothly rather than tearing, retaining letter legibility.
\textbf{\textit{Glyphs animate in Acrobat.}}}
\label{fig:method_teaser}
\end{figure}

Kinetic typography, text that moves, is used to convey tone, emphasis, and
narrative. In practice, it is almost always authored by hand, keyframe by keyframe, even with dedicated
tools for animating expressive text \citep{lee2002kinetic}. Generative models have recently started to automate the creative process, 
first by reshaping a letter toward a concept and leaving it static, then animating it under a video prior \citep{iluz2023word,tanveer2023ds,liu2025dynamic,park2024kinety}.
The main difficulty with automation is that the motion must satisfy three demands, two of which pull against each other. Specifically, motion should be \emph{expressive}:
the animation of a word should evoke its meaning, so that a reader feels the
snake in ``snake'' and the energy for a corresponding prompt that may say ``jump,'' for example. Yet the glyph must
remain \emph{legible}: deforming a glyph too freely could make it stop being a letter. In addition, the motion must
carry meaning of its own rather than ride on the shape. In the limit where the glyph may not be reshaped at
all, the action is the only element left to express the concept, which is referred to as the ``frozen-shape'' setting (see
\S\ref{sec:experiments}).

Recent work on Dynamic Typography \citep{liu2025dynamic} optimizes pixels or vertices directly against a generative prior,
where a single representation performs the two jobs: the very same boundary points must
move to sculpt the shape \emph{and} to animate it. Shape and motion then compete for the
same degrees of freedom, and the deformation that serves one is spent at the
expense of the other, which is why free-form optimization struggles to embed motion cleanly. 
At the other end of the spectrum, methods that impose structure, such as detected keypoints or per-group homographies, borrow it from category-specific priors (e.g., human skeletons) and generalize
poorly \citep{wu2025aniclipart,levy2026livesvg}. 
These two families of methods thus struggle for the same underlying reason, the \emph{representation}: what a method allows to move. Underneath it lies a second difficulty. The video prior speaks through score distillation (SDS), whose gradients are noisy from step to step, and whatever the representation leaves free, that noise is free to move: when each point and frame can move independently, the outline tears and the motion jitters as a result.

In this work, we address both difficulties by changing the \emph{representation}. 
Our key idea is to build motion from a glyph's \emph{natural vibration modes}. Viewed as a thin elastic sheet, a glyph outline has a small repertoire of ways it bends most easily, given by the \emph{eigenmodes} of its elasticity problem, which have long been used in graphics as a subspace for deformation
\citep{pentland1989good,barbivc2005real}. 
The softest of these eigenmodes span whole strokes and follow a letter's geometry, bending it smoothly rather than creasing it, e.g., 
a `V' opens and closes like a pair of wings, a `Q' swings its tail out from the bowl, and a `Y' spreads its arms. Only the first few modes 
are useful: stiffer modes bend the outline over ever shorter lengths until they crease it with legibility compromised. 
Figure~\ref{fig:method_teaser} shows three whole-letter modes for each example.
Restricting motion to this basis serves two goals at once: (a) natural motion, since the letter moves only in the ways it bends most 
easily, and (b) legibility, since the stiffer modes that would crease the outline are excluded.

Bending, however, is not the whole repertoire. The same eigenproblem has three zero-energy solutions: rigid translation in $x$
and $y$, plus rotation. When applied to a whole letter, these would only shift or turn it, but when confined to a part, they are 
exactly what a limb needs, letting the part move as a block while the joint absorbs the difference.
We therefore write these three modes in closed form for each part, with parts taken from a list generated once for the whole alphabet (\S\ref{sec:local}): 
the letter's mechanics decide how a part bends, and the list decides only where one part ends and the next begins. A frozen 
text-to-video model then only has to supervise the modes' amplitudes and phases over time (see purple arrow in Figure~\ref{fig:method_teaser}), 
a problem with a few dozen degrees of freedom rather than the thousands in free-form point optimization.

We address the noise in SDS gradients by pairing the modal basis above with a periodic drive, which expresses 
each mode's coefficient over time as a sum of a few whole-cycle sinusoids. When every point in every frame has 
its own parameters, this noise reaches the motion in two ways: in space, neighbouring points are pushed apart and 
the outline tears; in time, a point jumps between frames and the loop jitters. Our modes suppress tearing: each is 
a smooth field over the outline, so that neighbouring points move together, except across part joints, where an 
edge-length term penalizes stretching. At the same time, the drive suppresses jitter, where one set of sinusoid 
parameters shapes the whole loop, leaving no per-frame freedom for noise to occupy. SDS can therefore move the 
animation only along directions that are smooth along the outline and repeat over the loop, and the same loss that 
produces jitter under a free per-frame drive yields smooth motion here (\S\ref{sec:ablations}, Appendix~\ref{sec:bandlimit}).

Our contributions can be summarized as follows:

\begin{itemize}

  \item \textbf{A modal motion basis computed from a glyph's outline in vector form}: a finite-element eigensolve over the 
  glyph's interior gives the softest ways the whole letter and each of its parts bend, and closed-form rigid modes let 
  each part move as a block. The layers are orthogonalized, so parts move as blocks rather than being bent into place, 
  and the basis displaces the outline itself, so every frame remains a resolution-independent vector path.

  \item \textbf{Smooth, looping motion intrinsic to the representation, not imposed by a dedicated loss}: a periodic drive that expresses 
  each mode's coefficient as a few whole-cycle sinusoids shared by all frames. Under the same SDS loss that makes the motion 
  jitter when each frame's coefficients are free, it yields smooth motion that loops seamlessly.
 
  \item \textbf{Shape-motion disentanglement by construction}: shape and motion live in separate representations, so they do 
  not compete for the same degrees of freedom. Motion cannot alter the base outline, and a letter can be animated without being reshaped.
  

\end{itemize}


  
\section{Related Work}
\label{sec:related}

\noindent
\textbf{Semantic and kinetic typography:} 
Word-as-Image \citep{iluz2023word} deforms a vector outline toward a concept under an 
image diffusion prior while regularizing legibility. DS-Fusion \citep{tanveer2023ds} blends glyph and 
concept in raster space via a discriminator-guided latent diffusion model. VitaGlyph 
\citep{feng2026vitaglyph} transforms only the part of a character that carries the concept, while 
MetaDesigner \citep{he2025metadesigner} coordinates language-model agents for glyph and texture. 
All four produce static results. 
Among animated approaches, Differentiable Variable Fonts 
\citep{parikh2026differentiable} make a variable font's design axes differentiable, enabling 
gradient-based editing and physics-based text animation, but stay within the variants the typeface defines. 
KineTy \citep{park2024kinety} trains a video diffusion model on kinetic-typography templates and outputs 
raster video, so its motion comes from templates rather than from the letter. 
Closest to our setting, Dynamic Typography \citep{liu2025dynamic} animates vector letters under a video 
diffusion prior by predicting per-frame control-point displacements with a neural displacement field, so its 
motion lives in a high-dimensional, per-frame parameterization. Our work shares the vector and prior-guided 
setting, but replaces free-vertex motion with a glyph-derived modal basis.

\vspace{2pt}

\noindent
\textbf{Score distillation and structured motion:} 
Score Distillation Sampling (SDS) \citep{poole2022dreamfusion} distills a pretrained diffusion model into the parameters of a differentiable renderer.
Like prior vector animation methods \citep{gal2024livesketch,liu2025dynamic}, we apply its video form through a differentiable vector rasterizer \citep{li2020differentiable}. 
When applied to per-frame vertex displacements, however, video SDS tends to tear outlines and jitter, so prior work stabilizes it by reducing the degrees of freedom 
through borrowed structures.
AniClipart \citep{wu2025aniclipart} moves keypoints from a pretrained detector with category templates under an as-rigid-as-possible deformation. 
Articulated Kinematics Distillation \citep{li2025articulated} distills motion onto a skeleton, while
Animus3D \citep{sun2025animus3d} and Sketch2Motion \citep{rai2026sketch2motion} use score distillation to animate 3D assets, with the latter through an inferred skeleton.
APAP \citep{yoo2024plausible} deforms a mesh through user handles under image SDS. 
LiveSVG \citep{levy2026livesvg} abandons distillation altogether, citing noisy gradients, and instead fits the SVG to a video sampled from an image-to-video model with per-group homographies. 
We observe that tearing and jitter stem largely from \emph{what} is optimized, hence we change the latter while keeping distillation. 
Specifically, the spatial basis comes from a glyph's own 
vibration modes rather than a rig, detected keypoints, or a sampled video, and a periodic drive shared by all frames yields smooth motion where a free per-frame drive under the same 
loss would lead to jitter (\S\ref{sec:ablations}).

\vspace{2pt}

\noindent
\textbf{Modal analysis and reduced deformable models:} 
Using natural vibration modes as a motion subspace is classical in graphics \citep{pentland1989good,barbivc2005real}, where 
a few low-frequency modes span the deformations a shape most easily undergoes. 
Modal warping \citep{choi2005modal} keeps such a subspace valid under large rotations, \citet{hauser2003interactive} drive 
constrained deformation interactively within it, and localized variants confine modes to a region, learned from examples \citep{neumann2013sparse} or solved on the geometry as Localized Manifold Harmonics \citep{melzi2018localized}. 
When geometry is unknown, projections of an object's vibration modes can instead be recovered from the temporal spectra of optical flow in video and used to simulate its response to new forces \citep{davis2015image}, and Generative Image Dynamics \citep{li2024gid} predicts such spectral motion from a single image with a diffusion model. However, both methods operate on raster pixels and derive motion from video, whereas in our work, a glyph's vector outline gives us its geometry, so we solve for its modes directly to displace the outline.

Reduced subspaces have also been used to control motion, e.g., optimal control \citep{barbivc2009deformable}, spacetime constraints \citep{hildebrandt2012interactive}, rest-shape actuation \citep{coros2012deformable}, and locomotion along a body's natural modes for legged characters \citep{kry2009modal,nunes2012using} and soft bodies \citep{benchekroun2024actuators}, with periodic oscillation long an animation primitive \citep{kass2008animating}. 
These control methods integrate motion from equations of motion or search for it against a hand-specified objective. 
In contrast, we prescribe the temporal coefficients analytically and let a frozen video prior supervise only their amplitudes and phases, so the motion is named in words rather than designed. 
On the basis itself, we confine modes to parts from a list generated once per alphabet, add closed-form rigid modes per part, and orthogonalize the layers, so a part's bending does not repeat what its rigid motion already expresses.

\section{Method}
\label{sec:method}

\begin{figure}[!t]
\centering
\includegraphics[width=\textwidth,trim=0 12 0 0,clip]{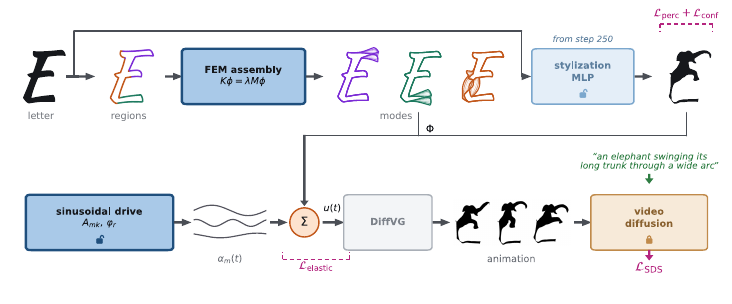}
\caption{\textbf{Pipeline of Modal Kinetic Typography:} The letter is partitioned into articulation parts from a fixed
per-alphabet table (Appendix~\ref{sec:parts}). Each part contributes three rigid modes, written in
closed form, and its two softest localized eigenmodes, alongside one whole-letter
mode, so the basis $\Phi$ is the letter's own. An analytic
integer-harmonic drive supplies the coefficients $\alpha_m(t)$; with $\Phi$ they give the
displacement $u(t)$, which rides on the sculpted template and is rendered differentiably. Sculpting
joins only at step $250$, since a shape that satisfies the prior early may conflict with the demand
that drives the motion. The frozen video prior sets how much of each mode is played, never what the
modes are. The basis is solved once on the source glyph and never recomputed. Open
padlocks mark the two optimized groups, the stylization MLP and the drive's $A_{mk},\varphi_r$.}
\label{fig:pipeline}
\end{figure}

Given a vector letter glyph and a concept caption, we produce a looping animation of the sculpted
glyph, deforming its outline over time. Every output frame is a deformed path, so the result is
resolution-independent vector graphics. Two groups of parameters are optimized and they stay
separate throughout. The first sculpts the glyph outline toward the concept. The second group sets only
how that outline moves over the loop, through a handful of coefficients shared by every frame.
Neither can do the other's work. Figure~\ref{fig:pipeline}
shows the full sequence and illustrates which loss controls which aspect of the pipeline.

\subsection{Shape sculpt: generating the base shape}
\label{sec:shape}
The glyph's shape is deformed toward the concept using the same machinery as Dynamic Typography
\citep{liu2025dynamic}. A small network offsets the outline control points under video SDS through a
differentiable rasterizer, with perceptual and conformal anchors preserving the source letter.
We, however, differ in two ways. First, motion is established before sculpting begins
(\S\ref{sec:loss}), so shape optimization occurs on an already moving glyph, a design set through empirical tests. Motion therefore
acts as a soft constraint so that edits that disrupt the learned articulation are discouraged. Second, we sculpt a single template rather than per-frame
geometry. Dynamic Typography can encode part of the concept in frame-specific offsets, which lets the form
drift from frame to frame. Our modal
drive animates one base outline, requiring the concept to reside fully in the base shape itself.

\subsection{Motion: a basis from the glyph's own mechanics}
\subsubsection{From glyph to motion basis}
\label{sec:mesh}
\label{sec:modes}
A mode is a shape the outline takes when the glyph flexes in a way its own geometry is soft to
move. We compute them at two levels, the whole letter and individual parts. Modes should act on large regions
rather than fine boundary detail, so we use a \emph{coarse} boundary,
$48$ uniformly spaced points, smoothed with a periodic Gaussian. Its interior is filled
with a constrained Delaunay quality mesh of $N$ vertices $V\in\mathbb{R}^{N\times2}$ and triangles
$T$. Modes solved on this mesh are interpolated back to the full-resolution glyph outline. Assembling linear (P1) element matrices over $T$, where the strain is constant within each element,
yields the global plane-stress stiffness matrix $K$ and the lumped mass matrix $M$
(Appendix~\ref{sec:impl}). Here $K$ is the Hessian of the elastic energy at rest, while $M$ accounts
for the amount of material that must be accelerated. The interior mesh is scaffolding and it only serves to assemble $K$ and $M$.
From $K$ and $M$ we then obtain the shape's vibration modes $\phi\in\mathbb{R}^{2N}$, one displacement per vertex, by solving the
generalized eigenproblem
\begin{equation}
K\phi \;=\; \lambda\, M\phi .
\label{eq:eig}
\end{equation}
The first three eigenvalues are zero and represent rigid translation in $x$ and $y$ and rotation, as they store no
strain energy. They are discarded from the whole-letter layer \emph{only}. \textbf{Rigid motion
is kept per part.} Each part receives its own three closed-form rigid modes, translation in $x$
and $y$ and rotation about the part's centroid, so a limb can pivot
without bending (\S\ref{sec:local} defines the parts, \S\ref{sec:hier} assembles the layers).
Among the remaining modes we keep the softest few. Since $\lambda = \phi^\top K\phi / \phi^\top
M\phi$ is elastic energy per unit of mass-weighted motion, a small $\lambda$ gives the most motion
for the least deformation, which means large, smooth bends and sways. Stiffer modes bend the
outline over ever shorter lengths and crease it, so they are dropped. Doubling the number kept per part buys no additional articulation (see the four-eigenmodes ablation, Table~\ref{tab:ablation}). The whole-letter layer of
the reported basis keeps only the single softest mode, as it behaved most consistently across letters
and typefaces in our runs, and widening this layer is left to future work. 
\subsubsection{Localized modes for parts}
\label{sec:local}
Global modes move the whole letter together. A part that articulates on its own needs modes confined to its region. Following Localized Manifold Harmonics \citep{melzi2018localized}, we tie every point \emph{outside} a region $r$ to its rest position with a stiff spring,
\begin{equation}
\big(K + \mu\,\mathrm{diag}(\mathbf{1}_{\notin r})\big)\,\phi \;=\; \lambda\, M\phi ,
\label{eq:lmh}
\end{equation}
so the softest modes have nowhere cheap to move but inside $r$. The mesh, $K$ and $M$ are shared across regions, only the mask changes per solve. The penalty scales with the stiffness itself ($\mu=10$ times the mean diagonal of $K$). Where the part table defines a joint band, the hard indicator is replaced by $1-w_r$ ($w_r \in [0,1]$) where $w_r{=}1$ inside the part and $w_r{=}0$ outside, with a smooth
transition across the band, so a limb can pivot about the joint rather than tear across it. Since
$\mu$ is finite, confinement is soft, so a mode can extend slightly beyond its region.

\phead{Defining regions:} Cutting the medial axis at its branch points was the obvious choice. However it returns a single region on unbranched letters like \textsf{C}, \textsf{I}, and \textsf{O}, so no part can move against another. We instead define the parts once per alphabet: a language model names each letter's moving parts and places one seed per part. This part table is shared by every glyph, typeface and caption, and solves the issue of unbranched letters. At run time seeds snap to the medial axis, the axis is partitioned by growing outward from them, and each boundary point takes the label of its nearest axis point. The parts are identical on every run. This approach captures geometrical intuition which a skeleton cannot, e.g., an \textsf{L} is one pen-stroke but
moves as two parts, an upright $|$ and a base $\_$.

\subsubsection{Hierarchical basis}
\label{sec:hier}
The basis has two layers: a \textbf{global} layer holding the single softest whole-letter mode
(\S\ref{sec:modes}), and a \textbf{part} layer holding each part's three rigid modes and two softest
localized eigenmodes (\S\ref{sec:local}). A letter of two, three or four parts therefore carries
$11$, $16$ or $21$ modes. The rigid modes are written in closed form
rather than solved for, because with the rest of the letter pinned the localized solver returns
swing mixed with bending rather than a clean pivot. The
whole-letter rigid modes are not kept to discourage the model from prioritizing global motion. Since the parts cover the whole letter, moving it requires every part's rigid modes to agree, keeping the option but not making it a short-cut. Orthogonalization keeps the modes separate: each mode is projected off the span of all modes
before it. Rigid modes are ordered first, therefore the eigenmodes lose the swing the rigid modes already express and carry only the
bending a pivot cannot. This does not change which deformations are reachable, but without it over
$80\%$ of an eigenmode's energy on average repeats earlier modes, mostly its part's rigid motion,
and the optimizer tends to bend parts into place instead of pivoting them. Two eigenmodes are kept per
part, and doubling that only doubles the drive spent on eigenmodes without adding motion
(Table~\ref{tab:ablation}, energy split in Appendix~\ref{sec:eigshare}). The whole basis is solved once on the source glyph,
before any sculpting, and held fixed for the run (Appendix~\ref{sec:impl}), so the motion stays the
letter's own even as the outline moves toward the concept.

\subsubsection{Motion model and periodic drive}
\label{sec:motion}
\label{sec:drive}
Let $\{\Phi_m\}$ be the assembled basis, all layers stacked as columns. The vertex displacement
field is
\begin{equation}
u(t) \;=\; \sum_m \alpha_m(t)\,\Phi_m .
\label{eq:motion}
\end{equation}
Here $\Phi_m$ is the displacement pattern of mode $m$ over the outline, and $\alpha_m(t)$ is a
scalar saying how much of that pattern is applied at time $t$. Its magnitude sets how far the
outline moves along the mode, and its sign sets the direction. Each frame is the base outline plus
$u(t)$, with $t$ sampled at $16$ evenly spaced points over one loop, so animating the letter
amounts to choosing the curves $\alpha_m(t)$.

Learning each coefficient freely at every frame leaves nothing tying consecutive frames together,
and the animation jitters (Table~\ref{tab:ablation}). We instead write each coefficient as a sum of
integer harmonics with a soft amplitude clamp,
\begin{equation}
\alpha_m(t) \;=\; a_{\max}\tanh\!\Big(\tfrac{1}{a_{\max}\sqrt{H}}\textstyle\sum_{k=1}^{H}
A_{mk}\,\sin(2\pi k\, t + \varphi_{r(m)})\Big), \qquad t\in[0,1).
\label{eq:drive}
\end{equation}
Here $a_{\max}$ is a fixed bound shared by all modes; $\tanh$ softly clamps each coefficient to it.
Animating with vibration modes and periodic oscillation has precedent in graphics
\citep{pentland1989good,kass2008animating,kry2009modal,nunes2012using,benchekroun2024actuators}.
We aim for looping animations typical of kinetic typography, so the frequencies are fixed to whole cycles of the clip, $k=1,\dots,H$ with $H{=}3$; the fundamental itself is fixed (Appendix~\ref{sec:speed}). The sum is normalized by $\sqrt{H}$ so the number of harmonics does not change the motion's magnitude. Every mode of a part shares \emph{one} phase $\varphi_{r(m)}$,
its rigid modes and eigenmodes alike, and the whole-letter mode has a phase of its own. This keeps the model simple, with a
single timing per part. What is learned is the relative phase between parts, which is the gait. The
amplitudes $A_{mk}$ are learned, one per mode and harmonic. A letter of three parts carries $16$
modes, so $48$ amplitudes and four phases, $52$ numbers for the entire animation. Smoothness then follows by construction, with every coefficient a sum of a few sinusoids shared by
all frames. There is no per-frame freedom for jitter to occupy, and each frame's gradient informs
every parameter rather than only its own (Appendix~\ref{sec:bandlimit}). Three harmonics per mode and a phase per part still leave
a wide range, from a single sway to parts moving at different rates and in counterpoint.

\subsubsection{Rendering, objective, and schedules}
\label{sec:loss}
The deformed paths are rasterized with DiffVG \citep{li2020differentiable}, and a frozen
text-to-video diffusion model supplies gradients through video SDS. Two anchors based on Dynamic
Typography \citep{liu2025dynamic}, perceptual and conformal, hold legibility to the source glyph, and
an edge-length term guards the part joints. Only the drive's amplitudes $A$ and phases $\varphi$ and
the shape network's weights are optimized, never vertex trajectories. The drive starts first and
sculpting joins at step $250$ (\S\ref{sec:shape}). Schedules are given in Appendix~\ref{sec:impl}.

\begin{figure}[!t]
\centering
\animategraphics[width=\textwidth,autoplay,loop,poster=first]{8}{figs/results_anim/frame_}{0}{11}
\caption{\textbf{Glyphs animated by their resonance:} Each row shows the input glyph followed by animation frames at fixed scale and position, beginning with the frame closest to the input. Motion combines the glyph's finite-element vibration modes, preserving smoothness and legibility. Shown are an \textsf{X} breakdancer and a strutting \textsf{R} rooster. In the frozen-shape setting (last row), the unsculpted \textsf{B} jiggles like jelly while retaining its original shape.}
\label{fig:teaser}
\end{figure}

\section{Experiments}
\label{sec:experiments}

We evaluate single-letter animation in two settings, against two published baselines and (as a
reference point rather than a baseline) Astra (GPT-6), a closed-source system that authors SVGs
directly. In the
\emph{full task} a method receives a letter and a caption and must both reshape the glyph
toward the concept and animate it, the setting Dynamic Typography \citep{liu2025dynamic}
addresses. In the \emph{frozen-shape} setting, the sculpting stage is switched off and the
method may only animate the glyph, which is where AniClipart \citep{wu2025aniclipart} is
directly comparable. Astra shares neither method's machinery: in both settings it receives our caption with the
letter and typeface to apply it to, and answers in one shot. Each test case is a letter, a typeface and a caption naming a
subject and its action, and produces a sixteen-frame animation. We use a language model to generate $100$ cases per setting with random letters, over
twelve typefaces. The two published baselines are evaluated at their
default setting.

\phead{Training Details:}
For the video diffusion model we use ModelScope \texttt{text-to-video-ms-1.7b} \citep{wang2023modelscope}, the same as published baselines, on a single $24$\,GB
GPU. Results are reported after $1000$ optimization steps, of which the first $250$ are motion
only. The basis composition and the learning-rate schedules are given in
Appendix~\ref{sec:impl}.

\phead{Metrics:}
\emph{Global} $\downarrow$ measures the glyph's overall movement and rotation and
\emph{Articulated} $\uparrow$ the remaining deformation, both as percentages of glyph size.
\emph{Jerk} $\downarrow$ is the third time-derivative of position over mean speed, and
\emph{Dead time} $\downarrow$ its counterweight, the share of samples whose acceleration is
negligible. \emph{Stretch} $\downarrow$ is our tearing measure. \emph{Temporal consistency}
$\uparrow$ \citep{wu2025aniclipart} and \emph{Concept} $\uparrow$, an X-CLIP text--video score
\citep{liu2025dynamic}, come from the two baselines. Details in Appendix~\ref{sec:metrics}.

These metrics should not be interpreted in isolation since jerk, temporal consistency and stretch are all
flattered by a clip that does not move, and concept similarity can be won by a drawing with no letter in it. The
last column therefore asks for two properties at once. \emph{Moves \& legible (M\,\&\,L)} is the share whose
articulated motion is at least half a stroke width \emph{and} whose form still reads as its letter
in most frames. Beyond small motion, our margin there grows as the bar rises (Appendix~\ref{sec:tradeoff}).

\begin{table}[!t]
\centering
\caption{\textbf{Quantitative comparison and user study} in the
\textit{Full task} and \textit{Frozen shape} settings. The last two columns report user-study rankings:
first, second, and third place receive $2$, $1$, and $0$ points, respectively. The visuals question is omitted for
\textit{Frozen shape}.
}
\label{tab:main}
\footnotesize
\setlength{\tabcolsep}{1pt}
\begin{tabular*}{\textwidth}{@{\extracolsep{\fill}}lcccccccc|cc@{}}
\toprule
Method & Global$\,\downarrow$ & Artic.$\,\uparrow$ & Jerk$\,\downarrow$ & Dead$\,\downarrow$ & Temp.$\,\uparrow$ & Stretch$\,\downarrow$ & Conc.$\,\uparrow$ & M\,\&\,L$\,\uparrow$ & Motion$\,\uparrow$ & Visuals$\,\uparrow$ \\
\midrule
Dynamic Typography  & $7.23$ & $6.28$ & $2.67$ & $\mathbf{0.6\%}$ & $0.950$ & $9.05$ & $18.23$ & $14\%$ & $0.89$ & $0.82$ \\
Astra (Full task)  & $9.26$ & $\mathbf{12.40}$ & $\mathbf{0.25}$ & $49.1\%$ & $\mathbf{0.976}$ & $\mathbf{1.47}$ & $19.08$ & $12\%$ & $0.79$ & $0.90$ \\
Ours (Full task)  & $\mathbf{6.11}$ & $9.24$ & $0.84$ & $1.1\%$ & $0.966$ & $3.32$ & $\mathbf{19.54}$ & $\mathbf{27\%}$ & $\mathbf{1.32}$ & $\mathbf{1.28}$ \\
\midrule
AniClipart  & $9.14$ & $5.22$ & $1.58$ & $\mathbf{0.3\%}$ & $0.981$ & $1.24$ & $18.67$ & $19\%$ & $1.17$ & -- \\
Astra (Frozen shape)  & $\mathbf{2.49}$ & $2.03$ & $\mathbf{0.55}$ & $13.1\%$ & $\mathbf{0.998}$ & $\mathbf{1.11}$ & $18.35$ & $1\%$ & $0.34$ & -- \\
Ours (Frozen shape)  & $7.72$ & $\mathbf{10.83}$ & $0.92$ & $1.6\%$ & $0.966$ & $2.03$ & $\mathbf{18.86}$ & $\mathbf{43\%}$ & $\mathbf{1.48}$ & -- \\
\bottomrule
\end{tabular*}
\end{table}

\begin{figure}[!tb]
\centering
\animategraphics[width=\textwidth,autoplay,loop,poster=first]{8}{figs/compare_anim/frame_}{0}{11}
\caption{\textbf{Comparison with Dynamic Typography and Astra.} The left-most column of each block is the input
glyph, the prompt sits above, and the frames span the clip. Our motion stays within the letter's structure. The toe lifts while the boot shaft stays
upright, and the limbs swing while the \textsf{X} holds its crossing. Dynamic Typography either barely moves or deforms the glyph heavily to move at all, and
Astra draws a clean icon that keeps little of the letter.}
\label{fig:compare}
\end{figure}

\phead{Quantitative Comparison:} Against Dynamic Typography, we use a similar motion budget but
favor articulated deformation over rigid translation. Our stretch is less than half theirs and lower
in $95\%$ of cases, while concept similarity, temporal consistency, and motion smoothness are higher.

Against AniClipart, we roughly double articulation while translating the glyph less, from a
low-dimensional parameterization in both cases. AniClipart uses category-specific control directions from a pretrained
detector, ours are the glyph's own vibration modes. It leads on temporal consistency and stretch, partly due to its
as-rigid-as-possible objective and low-motion cases. Restricting to pairs
where both methods move, our articulation still leads ($13.44$ vs.\ $8.52$).

Astra leads several geometric and temporal metrics, but largely by construction. Its rigid parts and
analytic keyframe interpolation suppress jitter and tearing. This comes at the cost of glyph identity,
as it redraws the subject rather than deforming the letter. In the frozen-shape setting, where
redrawing is disallowed, its articulation collapses. Its high dead time reflects two failure modes:
$44\%$ of its ink never moves, and even across the half of the outline that moves most, $32\%$ of samples carry no acceleration. This is the signature of constant-velocity interpolation between keyframes which reads as less realistic.


\phead{User study:}
Each study contains $10$ cases, answered by $44$ participants on the full task and $39$ on the
frozen-shape task.
For each case, participants rank the three clips, scored $2$, $1$, and $0$ points and averaged in
Table~\ref{tab:main}'s last two columns (Appendix~\ref{sec:userstudy} gives the full protocol). The full task asks which clip moves best and which has the better visuals while keeping the letter's identity.
Frozen shape asks only about motion. For the full task, five cases are representative Dynamic Typography examples and five are sampled
from twenty cases where both Dynamic Typography and ours achieve strong concept alignment, all
regenerated by the three methods on the same letter, typeface and caption. Frozen-shape cases are
sampled from twenty where both AniClipart and ours exhibit strong motion. These criteria favor the baselines
by including only cases where they have strong outputs. Ours leads on both full-task questions and on
frozen-shape motion.


\phead{Qualitative Comparison:}
Our animations alone are in Figure~\ref{fig:teaser}. Figure~\ref{fig:compare} sets the full-task
ones beside Dynamic Typography and Astra, Figure~\ref{fig:aniclipart} against AniClipart and
Astra on frozen shape.
Against Dynamic Typography our motion is more articulated and smoother, and it is localized where
the caption asks for it rather than spread over the whole glyph. Against AniClipart it is again
more articulated, and it reads as the action rather than as a pose being nudged. Against Astra the
difference is what the motion belongs to. Ours deforms the letter it was given and stays that
letter, while Astra redraws the subject. In the frozen-shape setting, where it may not redraw,
it struggles to produce motion at all. The caption's own effect on the motion is shown in Appendix~\ref{sec:promptctl}.
\begin{figure}[!tb]
\centering
\animategraphics[width=\textwidth,autoplay,loop,poster=first]{8}{figs/aniclipart_anim/frame_}{0}{11}
\caption{\textbf{Comparison with AniClipart and Astra on the frozen-shape task.} All three animate
the same given outline, with no sculpting on any side. Driving the glyph's own modes gives a fuller
and more realistic articulation: the \textsf{S} undulates along its length and the \textsf{H} rocks
on its uprights. AniClipart's keypoint rig, whose control directions are the same for any clipart,
holds the letters closer to their rest pose, and Astra mostly declines to move the outline it is
handed.}
\label{fig:aniclipart}
\end{figure}


\section{Ablations}
\label{sec:ablations}

\phead{Free per-frame drive.} Each mode's coefficient is learned independently at every frame, in place of the harmonic drive of \S\ref{sec:drive}, with the basis and the loss unchanged. The animation jitters: jerk more than triples at the median and is worse in all $20$ pairs, and temporal consistency falls. As an alternative we also tried an MLP $\alpha=\mathrm{net}(t)$, zero-meaned to maintain looping. It collapsed to stillness, so we report the free table here and leave a more learnable $\alpha(t)$ to future work.

\phead{Global modes only.} Every mode is made a whole-letter eigenmode, keeping the mode count. Removing the part layer takes localization and per-part rigid motion together. Since the whole-letter rigid modes are discarded by design, this row bounds what the layer contributes rather than separating the two, which also explains its lower global score. Articulation is unchanged, but the motion concentrates at the free ends: \emph{Concentration}, the share of the total motion carried by the most-moving quarter of the outline, rises from $0.39$ to $0.48$. 

\phead{No orthogonalization.} Each layer's modes are kept as solved, rather than orthogonalized
against the layers before it. Part eigenmodes then repeat rigid motion (around $80\%$). Articulation is
unchanged, but parts favor bending over pivoting. The optimizer spends
$4\times$ as much drive on the eigenmodes ($0.22$ versus $0.05$, in all $20$ pairs). Global motion also goes up, with more
whole-letter drift in $14$ of $20$ pairs. The bird in Figure~\ref{fig:ablation} looks dragged rather than pivoted.

\phead{Four eigenmodes per part.} Doubling the basis doubles eigenmode drive ($0.10$ versus $0.05$) without a significant gain in articulation (higher in $9$ of $20$ pairs) or other metrics.

\begin{table}[!t]
\vspace{1pt}          
\centering
\caption{\textbf{Ablations} on the full task. Each row replaces one component, on the same $20$
cases, medians at step $1000$. In brackets, the paired Wilcoxon $p$ against the full model
on the same cases. No difference in Articulated is significant. Appendix~\ref{sec:ablfig} shows them qualitatively.}
\label{tab:ablation}
\footnotesize
\setlength{\tabcolsep}{2pt}
\begin{tabular*}{\columnwidth}{@{\extracolsep{\fill}}lcccccc@{}}
\toprule
Variant & Global $\downarrow$ & Artic.\ $\uparrow$ & \ifdefined\arxivbuild Concen.\else Conc.\fi\ $\downarrow$ & Jerk $\downarrow$ & Stretch $\downarrow$ & Temp.\ $\uparrow$ \\
\midrule
Full model           & $5.50$ & $8.92$ & $0.39$ & $0.78$ & $2.96$ & $0.96$ \\
\midrule
Free per-frame drive & $5.49$ ($.87$) & $9.96$ ($.65$) & $0.36$ ($.26$) & $2.86$ ($<.001$) & $4.65$ ($.11$) & $0.95$ ($.01$) \\
Global modes only & $1.65$ ($<.001$) & $8.84$ ($.78$) & $0.48$ ($<.001$) & $0.83$ ($.25$) & $2.78$ ($.39$) & $0.97$ ($.28$) \\
No orthogonalization & $6.38$ ($.04$) & $10.77$ ($.90$) & $0.38$ ($.41$) & $0.66$ ($.48$) & $3.20$ ($.55$) & $0.96$ ($.48$) \\
Four eigenmodes per part & $5.28$ ($.57$) & $11.32$ ($.73$) & $0.38$ ($.99$) & $0.76$ ($.73$) & $3.83$ ($.73$) & $0.96$ ($.62$) \\
\bottomrule
\end{tabular*}
\end{table}

\section{Conclusion, limitations, and future work}

We present modal kinetic typography, which builds a letter's motion from its natural vibration modes and per-part rigid modes, with a frozen video diffusion prior supervising only their amplitudes and phases. This yields smooth motion at less cost to legibility than free-form optimization and enables a letter to move without being reshaped. Since the smoothness comes from the representation rather than a dedicated loss, the modal approach should carry over to other SDS-driven dynamic content.

Our representation still has limits (Appendix~\ref{sec:limitations}). To start, a single outline cannot change topology or turn in depth, hence 
a glass cannot shatter, nor can a flame flicker away.
The drive is periodic by construction, excluding actions that end somewhere other than the start. More importantly,
our results depend on how well the letter and typeface suit the concept. In particular, parts come from a list fixed per letter rather than per concept, so when the subject's anatomy does not match the letter's parts, a limb may be forced to move with its neighbour or be split across two
parts. Separately, a part that does have its own region may still drift when the caption asks it
to stay put, such as the base of a hammer, since the supervision cues motion more strongly than its
absence. Finally, our part list covers the uppercase Latin alphabet only, and we evaluate single letters rather than words. 

Future work could add non-periodicity to the drive, explicit constraints for stationary parts, and support for other scripts and words, and automatically choose letters to best suit a prompt.

\clearpage
\bibliography{iclr2027_conference}
\bibliographystyle{iclr2027_conference}

\clearpage
\raggedbottom
\appendix

\section{Appendix}

\ifdefined\novenuestatements\else
\subsection{Reproducibility statement}
\label{sec:repro}
What the animation is allowed to do is fixed by construction. The modal basis is solved once on the
source glyph from the vector outline alone, with no learning and no per-instance rigging, and the
parts come from a table fixed per letter, so the basis and the partition are identical on every run
(\S\ref{sec:local}, Appendix~\ref{sec:parts}, which prints the table). What is sampled, and
therefore varies between runs, is the score distillation itself, since the video prior is drawn
fresh at each step. We therefore fix the seed, $0$ for every run reported here, and compare methods
across the whole case set rather than on single examples. Appendix~\ref{sec:impl}
gives the material constants, the solver setting and the learning-rate
schedules. \S\ref{sec:experiments} gives the video prior, the optimization budget and
the number of frames, and Appendix~\ref{sec:metrics} defines every measured column and the test each
comparison uses. The evaluation cases, that is the letter, typeface and caption of each case, were
generated once and are reused unchanged by every method. We will release the code and the
articulation-part table.

\subsection{Use of generative AI}
\label{sec:aiuse}
We disclose our use of generative AI tools (Claude and Gemini) following the ICLR policy for
authors. We used these tools to implement code, to debug it and to write test scripts, and they
also helped design the ablations and interpret the results. A language model generated the
evaluation case sets, that is the letter, typeface and caption of each case. Separately, a language
model authors the articulation-part table. They were further used to draft and improve the text for
readability, to format \LaTeX{}, to help make the figures and to search for a few additional
references to add to the related work. Finally, Astra
(GPT-6) appears in our experiments as a reference point we measure against
(\S\ref{sec:experiments}), which is a generative model evaluated as a subject of comparison rather
than one used to prepare this work. The authors take responsibility for the contents of this paper.

\fi

\subsection{Ablations, qualitatively}
\label{sec:ablfig}
\begin{figure}[H]
\centering
\animategraphics[width=\textwidth,autoplay,loop,poster=first]{8}{figs/ablation_anim/frame_}{0}{11}
\caption{\textbf{Ablations, qualitatively.} Each row compares the full model with an ablation using the same letter, caption, seed, and mode count. \emph{Top:} The free per-frame drive jitters between consecutive weightlifter frames. \emph{Middle:} Whole-letter modes move little beyond the hammock's rail tops. \emph{Bottom:} Without orthogonalization the bird is dragged rather than pivoted.}
\label{fig:ablation}
\end{figure}



\subsection{The caption controls the motion}
\label{sec:promptctl}
As the video prior acts only through modal coefficients, the caption controls \emph{how the
modes are played} rather than what they are. Figure~\ref{fig:prompt_control} holds the letter and
its typeface fixed and varies only the action. On \textsf{K}, ``raises his sword'' and ``swings his
sword'' give the same visual with distinct motions, where Dynamic
Typography~\citep{liu2025dynamic} produces nearly the same action for both.

\begin{figure}[H]
\centering
\animategraphics[width=\textwidth,autoplay,loop,poster=first]{8}{figs/prompt_anim/frame_}{0}{11}
\caption{One glyph, two captions for the same subject, all else fixed. Ours raises the sword in one and sweeps it down in the other, while Dynamic Typography~\citep{liu2025dynamic} moves similarly for both.}
\label{fig:prompt_control}
\end{figure}

\subsection{Speed is a knob}
\label{sec:speed}
Because the drive uses harmonics of one fundamental, action speed is set by a single value $f$ rather
than learned. Multiplying all harmonics by $f$ plays the basis $f$ times per clip.
Figure~\ref{fig:freq} shows $f\in\{0.5,1,2\}$ with everything else fixed: the lizard takes half,
one, or two strides in the same number of frames. The action is unchanged; only its rate varies. Integer $f$ preserves seamless looping because every harmonic completes an integer number of cycles, so $f{=}0.5$ breaks the loop.

\begin{figure}[H]
\centering
\animategraphics[width=\textwidth,autoplay,loop,poster=first]{8}{figs/freq_anim/frame_}{0}{11}
\caption{Effect of base frequency.}
\label{fig:freq}
\end{figure}

\subsection{User study details}
\label{sec:userstudy}
Participants were given the following brief, then shown one case per page. The brief describes the task as defined by prior work, Dynamic Typography~\citep{liu2025dynamic}. 
Both that method and ours deform the input outline, hence the brief's requirement applies to them equally.

\begin{quote}
Kinetic typography aims to \textbf{reshape a letter} toward a prompted \textbf{concept} and animate
it with \textbf{natural, temporally coherent, and semantically faithful motion}, while the result
stays recognizable as the letter. Both shape and motion should arise from the letter's own
structure, its strokes, parts, mass, and symmetry, rather than being pasted on, with motion that is
smooth, plausible, and timed to \textbf{convey the described action most naturally}.

Each page shows one prompt, the original letter, and the animations of it. Please watch every
animation fully, then rank them for each question. Give each rank to exactly one animation.
\end{quote}

\noindent The two questions are:

\begin{quote}
\textbf{Q1:} Which animation exhibits the most \textbf{realistic} and \textbf{smoothest} motion?

\textbf{Q2:} Which animation has most \textbf{realistic and high quality visuals} while keeping the
\textbf{letter's base identity}?
\end{quote}

\begin{figure}[!ht]
\centering
\includegraphics[width=\textwidth]{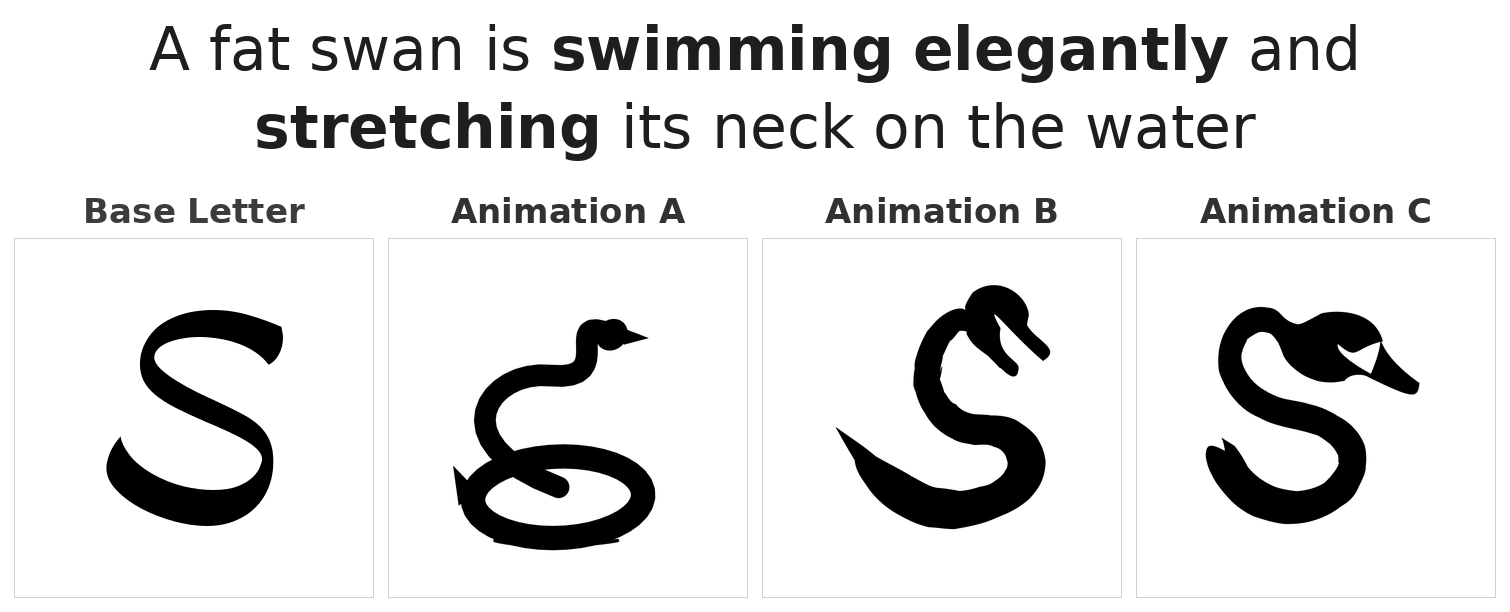}
\caption{\textbf{One page of the study.} The prompt, the base letter, and the three methods'
animations in a randomized order, labelled only A, B and C. Participants watched each clip in full
and then answered the two questions below the page, giving each rank to exactly one animation.}
\label{fig:userstudy_ui}
\end{figure}

Q1 is reported as \emph{Motion} in Table~\ref{tab:main} and Q2 as \emph{Visuals}; in the
frozen-shape setting, where no method may change the glyph, only Q1 is asked. Figure~\ref{fig:userstudy_ui} shows a page as participants saw it.

\subsection{Implementation details}
\label{sec:impl}
Operators use Poisson ratio $\nu{=}0.3$ and unit Young's modulus, which scales the eigenvalues
uniformly without changing mode shapes or their ranking. Eq.~\ref{eq:eig} is solved sparsely for the
smallest eigenpairs. The basis is solved once, on the source glyph before any sculpting, and is held
fixed for the whole run rather than being recomputed as the outline changes. Motion learning rates
warm up for $200$ steps to $0.30$ for amplitudes and $0.03$ for phases, then cosine-decay to half.
The shape learning rate warms up for $300$ steps, peaks at $1.1\times10^{-2}$, and log-linearly
decays to $4.4\times10^{-3}$. Every run uses seed $0$.

Each frame is rasterized differentiably at $320\times320$, and a clip is $16$ frames. Score
distillation uses classifier-free guidance at scale $30$, with the diffusion timestep drawn
uniformly from $[50, 949]$ of the model's $1000$-step schedule at every step and the usual
$(1-\bar\alpha_t)$ weighting on the gradient. The amplitude bound of Eq.~\ref{eq:drive} is
$a_{\max}=20.7$ px on that canvas, shared by every mode; modes are normalized to unit RMS
displacement, so the bound means the same thing for each of them. Of the loss terms in
\S\ref{sec:loss}, the perceptual and conformal anchors are weighted at $2000$ each and the
edge-length term at $100$.

\subsection{Visual example of modes}
\begin{figure}[!ht]
\centering
\includegraphics[width=\textwidth]{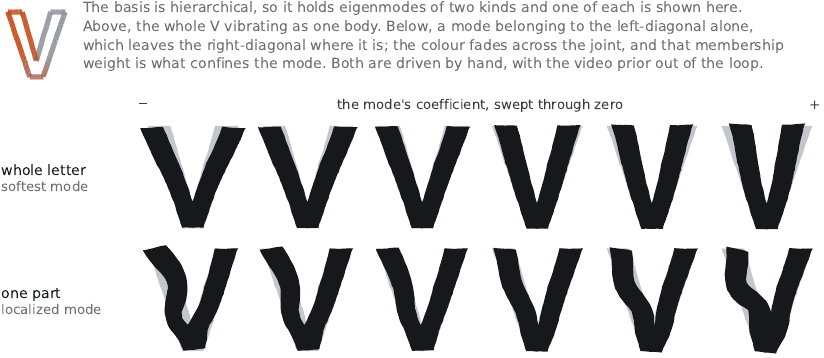}
\caption{\textbf{The two kinds of eigenmode, driven by hand.} A \textsf{V}: above, the softest
whole-letter mode; below, a mode confined to its left diagonal.}
\label{fig:modes}
\end{figure}

Figure~\ref{fig:modes} drives two modes of one letter by hand, without the video prior.

\subsection{Spread of energy across the basis}
\label{sec:eigshare}

\begin{table}[H]
\vspace{4pt}          
\centering
\caption{Mode for mode the eigenmodes carry a little more than
the closed-form rigid modes, and the optimizer drives both about equally hard.}
\label{tab:layers}
\small
\begin{tabular}{lcccc}
\toprule
Layer & Modes & Energy & Per mode & Peak coefficient \\
\midrule
Whole letter                    & $1$        & $8.2\%$  & $8.2\%$  & $7.72$ \\
Per part, rigid (closed form)   & $3$ each   & $53.0\%$ & $17.7\%$ & $6.15$--$6.60$ \\
Per part, localized eigenmodes  & $2$ each   & $38.8\%$ & $19.4\%$ & $6.52$--$7.43$ \\
\bottomrule
\end{tabular}
\end{table}

\subsection{What the drive can and cannot express}
\label{sec:bandlimit}
Score distillation may ask for anything, but only the part of its request that the basis can
express reaches the animation. In time the basis is three whole-cycle
harmonics per mode. Table~\ref{tab:bandlimit} gives the resulting motion spectrum, beside the free
per-frame drive of \S\ref{sec:ablations}, which changes only this. Above the third harmonic the full
model carries $0.0004$ of its moving power against the free table's $0.186$, higher in all $20$
cases, which is what the jerk column reports. 

\begin{table}[H]
\vspace{4pt}          
\centering
\caption{\textbf{What reaches the animation.} Share of moving power per temporal harmonic of the
realized motion, median over the $20$ ablation cases. The free per-frame drive keeps the same
spatial basis and changes only the temporal parameterization.}
\label{tab:bandlimit}
\small
\begin{tabular}{lccccc}
\toprule
Drive & $k{=}1$ & $k{=}2$ & $k{=}3$ & $k>3$ \\
\midrule
Full model (whole-cycle harmonics) & $0.598$ & $0.120$ & $0.081$ & $\mathbf{0.0004}$ \\
Free per-frame drive               & $0.495$ & $0.139$ & $0.055$ & $0.186$ \\
\bottomrule
\end{tabular}
\end{table}

\subsection{The articulation-part table}
\label{sec:parts}
\begin{figure}[!ht]
\centering
\includegraphics[width=0.96\textwidth]{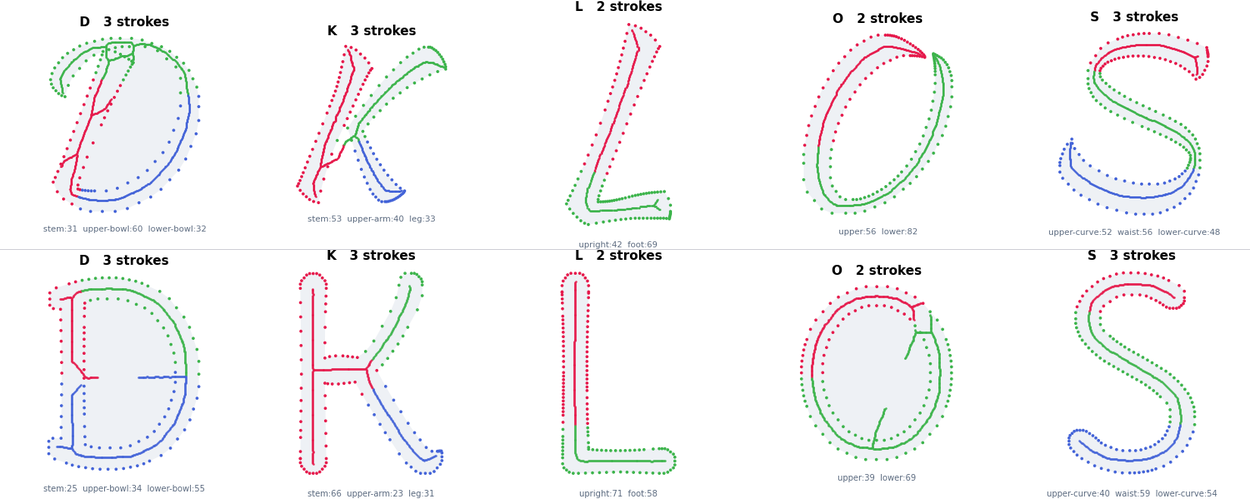}
\vspace{7pt}          
\caption{\textbf{The articulation-part table, and its transfer across typefaces.} Five letters
with their parts coloured, in two of the twelve typefaces we animate.}
\label{fig:parts_table}
\end{figure}

We show the table rather than describe it. Figure~\ref{fig:parts_table} gives the realized
partition for five letters, in two of the paper's twelve typefaces, chosen to be of very different
construction. The table has $26$ entries, or $75$ named parts, produced once for the alphabet.
Every case, concept and typeface in the paper reuses it unchanged. 

\subsection{Metrics and letter fidelity}
\label{sec:metrics}
Definitions of the measured columns of Table~\ref{tab:main}, in the order the table reports them,
together with the letter-match measure its last column is built on, followed by the shape measures
and the per-case evidence behind them. One measure is taken from each
published baseline so numbers stay comparable with the work they come from; the rest are ours.

\begin{itemize}
  \item \textbf{Motion}: reported as two separate numbers, because two quite different things both
  count as movement. A letter can \emph{travel}, sliding and turning across the frame while its form
  stays exactly as it was, and it can \emph{articulate}, with one part moving relative to the rest,
  which is what an action such as a leg swinging or a wing beating looks like. So for each
  frame we ask how much of the movement is explained by carrying the letter around as one rigid
  piece. That share is \emph{Global} $\downarrow$ and everything left over is
  \emph{Articulated} $\uparrow$, so the two never count the same movement twice. A glyph that
  merely flies across the canvas scores high on the first and near zero on the second.

  Concretely, every method logs the same $N$ outline points in each of the $T$ frames, so a clip is
  a table of positions rather than a video and the movement is read straight off it, instead of
  being estimated from the pixels, which is unreliable inside a solid silhouette. We take the
  letter's rest shape to be the average of its frames. For each frame we then find the single
  rotation and translation that best carries that rest shape onto it, in the least-squares sense,
  and apply it. Whatever this transform accounts for is the letter moving as one piece, the
  distance still separating the two shapes afterwards is movement no rigid motion can explain.
  Averaged over all points and frames, that leftover is \emph{Articulated}, and subtracting it from
  the total movement away from the rest shape leaves \emph{Global}. Both are divided by the
  letter's radius, the typical distance from its centre out to its outline, so the numbers read as
  percentages of letter size and compare across typefaces and canvas scales. Before the split we also trim
  the trajectory to the temporal frequencies carrying most of its movement, which sets a little of
  the frame-to-frame flicker aside.

  \item \textbf{Jerk} $\downarrow$: the third time-derivative of the control-point positions, divided by mean speed. Dividing by speed makes it scale-free, so a method is not penalised for moving further.

  \item \textbf{Dead time} $\downarrow$: the share of trajectory samples whose acceleration is negligible next to what a single loop-frequency harmonic of the same speed would have. The test is on acceleration alone, so coasting at constant velocity counts as dead exactly as standing still does. Jerk is minimised by motion that barely happens, so dead time is its counterweight.

  \item \textbf{Temporal consistency} $\uparrow$: the mean cosine similarity between consecutive-frame image features, as reported by \citet{wu2025aniclipart}. While designed to measure smooth motion, it is maximised by a glyph that does not move.

  \item \textbf{Stretch} $\downarrow$: $\max_i \ell_i/\ell^0_i$ over boundary edges relative to rest length, our tearing measure. Rest is the frame-mean shape, which for a zero-mean drive is the template itself. Edges shorter than a tenth of the median rest length are excluded, since dividing by them detects pinching rather than tearing, and no edge is ever formed between two contours.

  \item \textbf{Concept} $\uparrow$: X-CLIP score between the clip and the caption \citep{liu2025dynamic}. Shuffling frames alters the score by $<2\%$, so we treat it strictly as a concept-similarity measure and leave motion to the user study.

  \item \textbf{Reads as letter}: the share of frames that still read as the intended letter. Each frame's ink is binarized, cropped and scaled into a fixed box with aspect ratio preserved, then matched by intersection over union against all $26$ uppercase letters rendered in a bank of typefaces, keeping the best typeface per letter, so the question is whether the form reads as \emph{some} \textsf{A} rather than one specific \textsf{A}. The case's own typeface family is held out, so a method that deforms that outline earns no credit for having started from it; holding it out costs our score five points, Dynamic Typography's six and Astra's one. On clean glyphs set in a held-out typeface the matcher recovers the right letter $88\%$ of the time.

  \item \textbf{Moves \& legible} $\uparrow$ asks for a pair of properties at once. Every column above can be won without solving the task. Jerk, temporal consistency and stretch are all flattered by a clip that does not move, and concept similarity by a drawing with no letter in it or even no motion. This metric thus is the share of cases whose articulated motion is at least half a stroke width \emph{and} whose form still reads as its letter in most frames. The bar is set in stroke widths because that is the scale of the thing being moved, the median stroke width of the evaluation typefaces is $14.7\%$ of glyph size, so the bar is $7.4\%$. Beyond small motion our lead grows as the bar rises, and Figure~\ref{fig:tradeoff} sweeps it.
\end{itemize}

\subsection{Expressiveness against legibility}
\label{sec:tradeoff}
\begin{figure}[!ht]
\centering
\includegraphics[width=\textwidth]{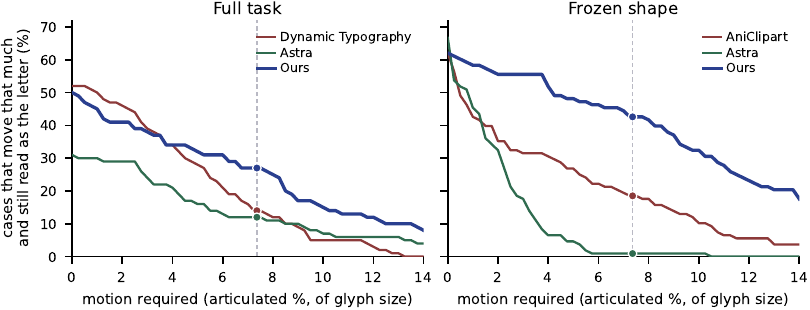}
\caption{\textbf{Motion against legibility.} The dashed line is the bar used for the
\emph{moves \& legible} column of Table~\ref{tab:main}.}
\label{fig:tradeoff}
\end{figure}

The tradeoff our representation is designed to weaken appears clearly across cases. As motion
increases, legibility tends to fall, especially for the baselines. The correlation is
$\rho=-0.30$ for Dynamic Typography and $-0.40$ for Astra, compared with $-0.21$ for ours.
Figure~\ref{fig:tradeoff} shows the same pattern directly. At modest motion, ours and Dynamic
Typography are similar ($50\%$ vs.\ $52\%$ legible), but the gap widens as more movement is
required. At half a stroke width, $27\%$ of our cases both move that much and remain legible,
compared with $14\%$ and $12\%$ for the baselines. In the frozen-shape setting, the same point is
$43\%$ for ours, versus $19\%$ and $1\%$. The two panels fall away for different reasons. On the
full task the baselines trade the letter for motion: among the clips that clear the bar, only
$44\%$ of Dynamic Typography's and $19\%$ of Astra's still read as their letter, against $47\%$
of ours. In the frozen-shape setting AniClipart instead rarely reaches that much motion at all,
clearing the bar in $32\%$ of cases against our $70\%$, while among the clips that do clear it
legibility is alike ($57\%$ against $61\%$). All curves begin below $100\%$ because the matcher,
with the case's own family held out, already misses the clean unanimated glyph in the harder
display and script faces. That floor is where the frozen-shape curves start together; on the full
task Astra starts lower still, since it redraws the letter rather than deforming it. Either way, our glyph-derived basis buys motion without spending the letter.

\subsection{Limitations}
\label{sec:limitations}
\begin{figure}[H]
\centering
\includegraphics[width=\textwidth]{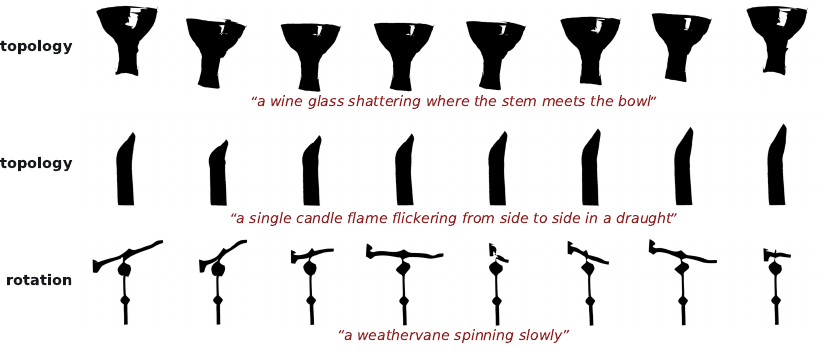}
\caption{\textbf{Three captions that ask for something outside the representation.} \emph{Top:}
the glass is asked to shatter, but one closed outline cannot separate, so it wobbles whole.
\emph{Middle:} the flame is asked to flicker, which would need tongues that detach and die away,
so the wick only leans. \emph{Bottom:} the vane is asked to turn, but the in-plane rigid mode is
linearized, so the arm rocks and returns rather than coming round. Eight frames spanning the clip
in each row.}
\label{fig:limits}
\end{figure}

\phead{What the representation cannot express.} Every frame is a displacement field on one base
outline, so the letter deforms without changing topology. Actions that break apart or dissipate, such as fire, are outside the method (Figure~\ref{fig:limits}). For the same reason a turn in depth is unavailable, since
there is no far side of the glyph to bring into view and no occlusion to reorder. Rotation within
the plane is available through each part's rigid rotation mode, but it is the linearized one, which
carries points along the tangent rather than the arc, so a part driven far around it lengthens
rather than turns. The animation is likewise a closed loop by construction, since
the drive is built from whole-cycle harmonics. We choose this for seamless repetition, at the cost
of actions meant to end somewhere other than where they began; nothing in the basis requires it, and
a non-returning component in the drive would lift the restriction.

\begin{figure}[H]
\centering
\includegraphics[width=\textwidth]{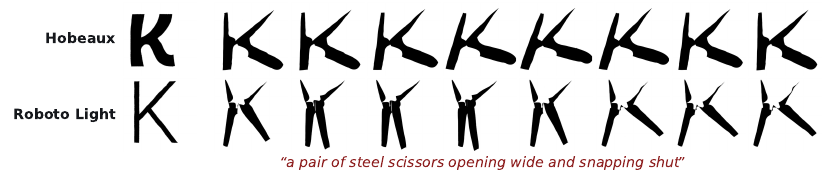}
\caption{\textbf{The same letter and caption in two typefaces.} Nothing differs but the face
the \textsf{K} is set in, and the seed is the same. The heavier face keeps thick blades that part
a little, the lighter one gives thin blades that open wide. }
\label{fig:fontdep}
\end{figure}

\phead{Sensitivity to the letter and the typeface.} The available motion is the letter's own, so
the outcome depends on how a glyph's geometry suits the concept, and the same concept on the same
letter animates differently in a different typeface. On \textsf{K} asked for ``a pair of steel
scissors opening wide and snapping shut'', a heavy face keeps thick blades that part a little
(articulation $8.0$) while a light one gives thin blades that open wide (articulation $17.0$), both plausible
scissors with different motion (Figure~\ref{fig:fontdep}).

\begin{figure}[H]
\centering
\includegraphics[width=0.82\textwidth]{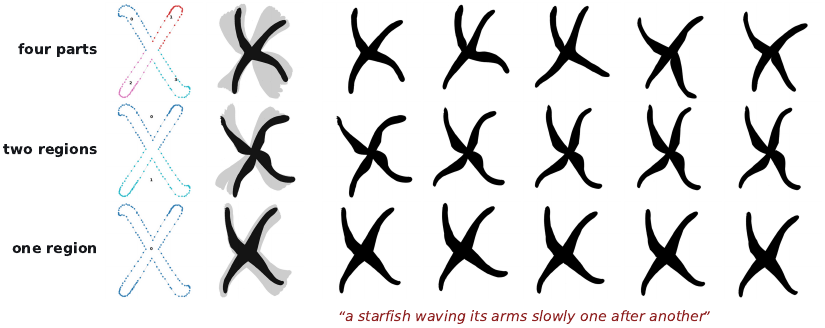}
\vspace{6pt}          
\caption{\textbf{How finely the glyph is divided decides how much of the motion is articulation.}
Same letter, caption, typeface and seed; only the partition changes. With the part table's four
parts each arm moves by its own amount, which is what the caption asks for ($16.8\%$ of glyph size,
rigid share $0.21$). Split into a top and a bottom half, the two arms in each half can only move
together ($8.7\%$, $0.35$), and as one region the letter can only lean as a whole ($3.4\%$, $0.67$).}
\label{fig:regions}
\end{figure}

\phead{The partition does not adapt to the concept.} The number of parts is itself controllable,
and the part table can name as many as a letter needs. What is fixed is that the decomposition is
decided once per letter, while the decomposition that suits the animation may
depend on the concept. An \textsf{X} could be a pair of scissors or a starfish: the scissors divide
naturally into two blades, whereas the starfish needs a finer division for each arm to move on its
own. Figure~\ref{fig:regions} shows the cost of too coarse a division, merging the starfish's four
parts into halves and then into one taking most of its articulation away. Future work could choose
the number of parts and their assignment from the target concept rather than from the letter
alone.

\begin{figure}[H]
\centering
\includegraphics[width=\textwidth]{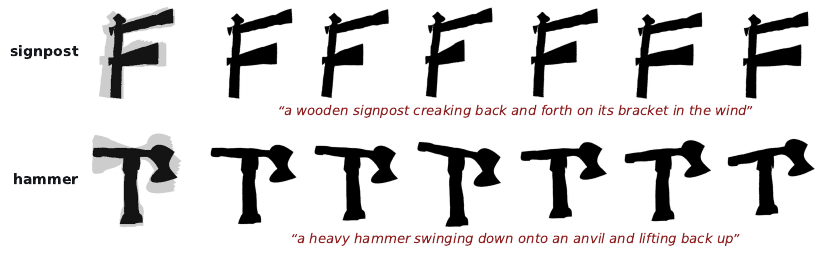}
\caption{\textbf{Parts the caption holds fixed still move.} The signpost's post moves $0.81$ times
as much as the sign it carries, and the hammer's base $0.96$ times as much as the head that swings,
where both should be near zero.}
\label{fig:stationary}
\end{figure}

\phead{Stationary parts and motion suppression.}
While our motion representation can assign different amplitudes to individual modes and parts, we observe that parts expected to remain stationary may still exhibit unintended motion. For example, an \textsf{F} signpost swings its sign but carries the post along with it, and a \textsf{T} hammer swings its head but carries its base, where both supports should stay near still (Figure~\ref{fig:stationary}). In principle, the optimization could suppress such motion by learning sufficiently small amplitudes for the corresponding components, suggesting that this behavior is not necessarily a limitation of the representation itself. Rather, the optimization may not always provide a sufficiently strong signal to distinguish between parts that should actively move and those that should remain fixed. One possible explanation is that the motion supervision provides stronger cues for the desired movement than for the absence of movement in stationary regions. Incorporating explicit stationary constraints or stronger spatial guidance could help address such cases.


\end{document}